\pdfoutput=1

\documentclass[11pt]{article}

\usepackage[]{acl}

\usepackage{times}
\usepackage{latexsym}

\usepackage{caption}
\usepackage{subcaption}

\usepackage[T1]{fontenc}

\usepackage[utf8]{inputenc}

\usepackage{microtype}

\newif\ifcomment\commentfalse

\usepackage[a-1b]{pdfx}

\usepackage{framed}
\usepackage{mdwlist}
\usepackage{latexsym}
\usepackage{colortbl}
\usepackage{xcolor}
\usepackage{nicefrac}
\usepackage{booktabs}
\usepackage{fnpct}
\usepackage{amsfonts}
\usepackage[T1]{fontenc}
\usepackage{bold-extra}
\usepackage{amsmath}
\usepackage{amssymb}
\usepackage{bm}
\usepackage{graphicx}
\usepackage{mathtools}
\usepackage{microtype}
\usepackage{multirow}
\usepackage{multicol}
\usepackage{todonotes}
\usepackage{latexsym,comment}
\usepackage[normalem]{ulem}

\newcommand*{\missingreference}{{\Huge \colorbox{red}{?reference?}}}
\newcommand*{\missingcitation}{{\Huge \colorbox{red}{?citation?}}}
\makeatletter
\def\@setref#1#2#3{%
    \ifx#1\relax
        \protect\G@refundefinedtrue
        \nfss@text{\reset@font\missingreference}%
        \@latex@warning{Reference `#3' on page \thepage \space
            undefined}%
    \else
        \expandafter#2#1\null
    \fi}
\def\@citex[#1]#2{\leavevmode
    \let\@citea\@empty
    \@cite{\@for\@citeb:=#2\do
        {\@citea\def\@citea{,\penalty\@m\ }%
            \edef\@citeb{\expandafter\@firstofone\@citeb\@empty}%
            \if@filesw\immediate\write\@auxout{\string\citation{\@citeb}}\fi
            \@ifundefined{b@\@citeb}{\hbox{\reset@font\missingcitation}%
                \G@refundefinedtrue
                \@latex@warning
                {Citation `\@citeb' on page \thepage \space undefined}}%
            {\@cite@ofmt{\csname b@\@citeb\endcsname}}}}{#1}}
\makeatother

\newcommand{\gem}[1]{\mbox{\textsc{gem}}}
\newcommand{\abr}[1]{\textsc{#1}}
\newcommand{\camelabr}[2]{{\small #1}{\textsc{#2}}}

\newcommand{\smallemaillink}[2]{{\small \href{mailto://#2}{\texttt{#1}}}}

\newcommand{\hidetext}[1]{}
\newcommand{\ignore}[1]{}

\ifcomment
    \newcommand{\pinaforecomment}[3]{\colorbox{#1}{\parbox{.8\linewidth}{#2: #3}}}
\else
    \newcommand{\pinaforecomment}[3]{}
\fi

\newcommand{\jbgcomment}[1]{\pinaforecomment{red}{JBG}{#1}}

\newcommand{\yfcomment}[1]{\pinaforecomment{brown}{YF}{#1}}

\newcommand{\smallurl}[1]{ \begin{tiny}\url{#1}\end{tiny}}

\definecolor{lightblue}{HTML}{3cc7ea}
\definecolor{CUgold}{HTML}{CFB87C}
\definecolor{grey}{rgb}{0.95,0.95,0.95}
\definecolor{ceil}{rgb}{0.57, 0.63, 0.81}
\definecolor{UMDred}{HTML}{ed1c24}
\definecolor{UMDyellow}{HTML}{ffc20e}

\newcommand{\conll}{\textsc{c}{\small o}\textsc{nll}}

\newcommand{\bert}{\abr{bert}}
\newcommand{\mbert}{\camelabr{m}{bert}}

\usepackage{siunitx}

\title{Match the Script, Adapt if Multilingual: Analyzing the Effect of Multilingual Pretraining on Cross-lingual Transferability}

\jbgcomment{Can we change the title?  If so, we might want to be more definitive about our findings.}

\author{Yoshinari Fujinuma\thanks{\hspace{0.1cm} This work was done while the first author was a student at University of Colorado Boulder.}\\
 \abr{aws} \abr{ai} Labs \\
  Amazon.com \\
  \smallemaillink{fujinumay@gmail.com} \\\And
  Jordan Boyd-Graber \\
  \abr{umiacs}, \abr{cs}, \abr{lsc}, iSchool \\
  University of Maryland \\
  \smallemaillink{jbg@umiacs.umd.edu} \\\And
  Katharina Kann \\
  Computer Science \\
  University of Colorado Boulder \\
  \smallemaillink{katharina.kann@colorado.edu} \\
  }

\begin{document}
\maketitle
\begin{abstract}
Pretrained multilingual models enable zero-shot learning even for unseen languages, and that performance can be further improved via adaptation prior to finetuning.
However, it is unclear how the number of pretraining languages influences a model's zero-shot learning for languages unseen during pretraining.
To fill this gap, we ask the following research questions: 
(1) How does the number of pretraining languages influence zero-shot performance on unseen target languages? (2) Does the answer to that question change with model adaptation? (3) Do the findings for our first question change if the languages used for pretraining are all related? 
Our experiments on pretraining with \emph{related} languages
indicate that choosing a diverse set of languages is crucial.
\emph{Without} model adaptation, surprisingly, increasing the number of pretraining languages yields better results up to adding related languages, after which performance plateaus.
In contrast, \emph{with} model adaptation via continued pretraining, pretraining on a larger number of languages often gives further improvement, suggesting that model adaptation is crucial to exploit additional pretraining languages.\footnote{All code used in this paper is available at \url{https://github.com/akkikiki/multilingual_zeroshot_analysis}.}
\end{abstract}

\section{Introduction}

Pretrained multilingual language models \cite{devlin2019bert,conneau2020xlmr} are now a standard approach for cross-lingual transfer in natural language processing (NLP).
However, there are multiple, potentially related issues on pretraining multilingual models.
\citet{conneau2020xlmr} find the ``curse of multilinguality'': 
for a fixed model size, zero-shot performance on target languages seen during pretraining increases with additional pretraining languages only until a certain point, after which performance decreases.
\citet{wang2020negative} also report ``negative interference'', where monolingual models achieve better results than multilingual models, both on subsets of high- and low-resource languages.
However, those findings are limited to target languages seen during pretraining.

Current multilingual models cover only a small subset of the world's languages. Furthermore, due to data sparsity, monolingual pretrained models are not likely to obtain good results for many low-resource languages. In those cases, multilingual models can  zero-shot learn for unseen languages with an above-chance performance, which can be further improved via model adaptation with target-language text~\citep{wang2020extending}, even for limited amounts~\citep{ebrahimi2021adapt}. However, it is poorly understood how the number of pretraining languages influences performance in those cases.
Does the ``curse of multilinguality'' or ``negative interference'' also 
impact performance on 
unseen target languages?
And, if we want a model to be applicable to as many unseen languages as possible, how many languages should it be trained on?






Specifically, we ask the following research questions: 
(1) How does pretraining on an increasing number of languages impact zero-shot performance on unseen target languages?
(2) Does the effect of the number of pretraining languages change with model adaptation to target languages?
(3) Does the answer to the first research question change if the pretraining languages are all related to each other? 
We pretrain a variety of monolingual and multilingual models, which we then finetune on English and apply to three zero-shot cross-lingual downstream tasks in unseen target languages: part-of-speech~(\abr{pos}) tagging, named entity recognition~(\abr{ner}), and natural language inference~(\abr{nli}).
Experimental results suggest that choosing a diverse set of pretraining languages is crucial for effective transfer.
Without model adaptation, increasing the number of pretraining languages improves accuracy on unrelated unseen target languages at first and plateaus thereafter. 
Last, with model adaptation, additional pretraining languages beyond English generally help. 

We are aware of the intense computational cost of pretraining and its environmental impact~\citep{strubell2019energy}. Thus,
our experiments in Section~\ref{sec:results} are on a relatively small scale with a fixed computational budget for each model and on relatively simple NLP tasks (\abr{pos} tagging, \abr{ner}, and \abr{nli}), but validate our most central findings in Section~\ref{sec:xlmr} on large publicly available pretrained models. 

\section{Cross-lingual Transfer via Pretraining}

Pretrained multilingual models are a straightforward cross-lingual transfer approach: a model pretrained on multiple languages is then fine-tuned on target-task data in the \emph{source} language. Subsequently, the model is applied to target-task data in the \emph{target} language. Most commonly, the target language is part of the model's pretraining data. However, cross-lingual transfer is possible even if this is not the case, though performance tends to be lower. This paper extends prior work exploring the cross-lingual transfer abilities of pretrained models for \emph{seen} target languages depending on the number of pretraining languages to \emph{unseen} target languages.
We now transfer via pretrained multilingual models and introduce the models and methods vetted in our experiments.

\subsection{Background and Methods}
\paragraph{Pretrained Language Models}
Contextual representations such as ELMo~\citep{peters2018elmo} and \bert{}~\citep{devlin2019bert} are not just useful for monolingual representations.  Multilingual \bert{}~\citep[\mbert{}]{devlin2019bert}, \abr{xlm}~\citep{conneau2019xlm}, and XLM-RoBERTa~\citep[\abr{xlm-r}]{conneau2020xlmr} have surprisingly high cross-lingual transfer performance compared to the previous best practice: static cross-lingual word embeddings~\citep{pires2019multilingual,wu2019beto}.
Multilingual models are also practical---why have hundreds of separate models for each language when you could do better with just one?
Furthermore, \citet{wu2020languages} report that models pretrained on 100+ languages are better than bilingual or monolingual language models in zero-shot cross-lingual transfer.

\paragraph{Model Adaptation to Unseen Languages}
Adapting pretrained multilingual models such as \mbert{} and \abr{xlm-r} to unseen languages is one way to use such models beyond the languages covered during pretraining time.
Several methods for adapting pretrained multilingual language models to unseen languages have been proposed, including continuing masked language model (\abr{mlm}) training~\citep{chau2020parsing,muller2020when}, optionally adding Adapter 
modules~\citep{pfeiffer2020mad}, or extending the vocabulary of the pretrained models~\citep{artetxe2020crosslingual,wang2020extending}. 
However, such adaptation methods assume the existence of sufficient monolingual corpora in the target languages.
Some spoken languages, dialects, or extinct languages lack monolingual corpora to conduct model adaptation, which motivates us to look into languages unseen during pretraining. 
We leave investigation on the effect of target language-specific processing, e.g., transliteration into Latin scripts~\citep{muller2021unseen}, for future work.

\jbgcomment{previous paragraph is a little disjointed, consider rewriting with a clear topic sentence}

\subsection{Research Questions}
A single pretrained model that can be applied to any language, including those unseen during pretraining, is both more efficient and more practical than pretraining one model per language. 
Moreover, it is the only practical option for unknown target languages or for languages without enough resources for pretraining.
Thus, models that can be applied or at least easily adapted to unseen languages are an important research focus. This work addresses the following research questions (\abr{rq}), using English as the source language for finetuning.

\noindent\textit{{\bf RQ1}: How does the number of pretraining languages influence zero-shot cross-lingual transfer of simple NLP tasks on unseen target languages?}

We first explore how many languages a model should be pretrained on
if the target language is unknown at test time or has too limited monolingual resources for model adaptation. 
On one hand, we hypothesize that increasing the number of pretraining languages will improve performance, as the model sees a more diverse set of scripts and linguistic phenomena. Also, the more pretraining languages, the better chance of having a related language to the target language.
However, multilingual training can cause interference:  other languages could distract from English, the finetuning source language, and thus, lower performance.

\noindent\textit{{\bf RQ2}: How does the answer to RQ1 change with model adaptation to the target language?}

This question is concerned with settings in which we have enough monolingual data to adapt a pretrained model to the target language.
Like our hypothesis for \abr{rq}1, we expect that having seen more pretraining languages should make adaptation to unseen target languages easier. However, another possibility is that adapting the model makes any languages other than the finetuning source language unnecessary; performance stays the same or decreases when adding more pretraining languages. 

\noindent\textit{{\bf RQ3}: Do the answers to RQ1 change if all pretraining languages are related to each other?}

We use a diverse set of pretraining languages when exploring \abr{rq}1, since we expect that to be maximally beneficial. 
However, the results might change depending on the exact languages.
Thus, as a case study, we repeat all experiments
using a set of closely related languages.
On the one hand, we hypothesize that benefits due to adding more pretraining languages (if any) will be smaller with related languages, as we reduce the diversity of linguistic phenomena in the pretraining data. However, on the other hand, if English is all we use during fine-tuning, performance might increase with related languages, as this will approximate training on more English data more closely.

\jbgcomment{Having more specific section titles is an easy way to punch up the paper}

\section{Experimental Setup}
\label{sec:experimental-setup}
\paragraph{Pretraining Corpora}
All our models are pretrained on the \conll{} 2017 Wikipedia dump~\citep{conll2017wikipedia}. 
To use equal amounts of data for all pretraining languages,
we downsample all Wikipedia datasets to an equal number  of sequences.
We standardize to the smallest corpus, Hindi.
The resulting pretraining corpus size is around 200MB per language.\footnote{\citet{micheli2020importance} show that corpora 
of at least 100MB are reasonable for pretraining.}
We hold out 1K sequences with around 512 tokens per sequence after preprocessing as a development set to track the models' performance during pretraining.


\paragraph{Corpora for Model Adaptation} 
For model adaptation (RQ2), we select unseen target languages contained in both \abr{xnli}~\citep{Conneau2018xnli} and Universal Dependencies 2.5~\citep{ud2.5}: 
Farsi (\textsc{fa}), Hebrew (\textsc{he}), French (\textsc{fr}), Vietnamese (\textsc{vi}), Tamil (\textsc{ta}), and Bulgarian (\textsc{bg}).
Model adaptation is typically done for low-resource languages not seen during pretraining because 
monolingual corpora are too small~\citep{wang2020extending}. 
Therefore, we use the Johns Hopkins University Bible corpus by \citet{mccarthy2020johns} following \citet{ebrahimi2021adapt}.\footnote{In cases where multiple versions of the Bible are available in the target language, we select the largest one.} 

\paragraph{Tasks}
We evaluate our pretrained models on the following downstream tasks from the \abr{xtreme} dataset~\citep{hu20icml}:
\abr{pos} tagging and \abr{nli}.
For the former, we select 29 languages from Universal Dependencies v2.5~\citep{ud2.5}. For the latter, we use all fifteen languages in  \abr{xnli}~\citep{Conneau2018xnli}.
We follow the default train, validation, and test split in  \abr{xtreme}.


\begin{table}[t!]
\centering
\small
\setlength{\tabcolsep}{10pt}
\begin{tabular}{lllllll}
\toprule
\bf Langs             & \bf Tasks  \\
\midrule
\bf Seen languages \\
\midrule
 English (\textsc{en})    & POS, NER, NLI\\
 Russian (\textsc{ru})    & POS, NER, NLI\\
 Arabic  (\textsc{ar})    & POS, NER, NLI\\
 Chinese (\textsc{zh})    & POS, NER, NLI\\
 Hindi   (\textsc{hi})    & POS, NER, NLI\\
 Spanish    (\textsc{es}) & POS, NER, NLI  \\
 Greek      (\textsc{el}) & POS, NER, NLI  \\
 Finnish    (\textsc{fi}) & POS, NER      \\
 Indonesian (\textsc{id}) & POS, NER      \\
 Turkish    (\textsc{tr}) & POS, NER, NLI  \\
 German  (\textsc{de})    & POS, NER, NLI\\
 Dutch   (\textsc{nl})    & POS, NER, NLI\\
 Swedish (\textsc{sv})    & - \\
 Danish  (\textsc{da})    & - \\
\midrule
\bf Unseen languages \\
\midrule
 Bulgarian  (\textsc{bg}) & POS, NER, NLI  \\
 French     (\textsc{fr}) & POS, NER, NLI  \\
 Urdu       (\textsc{ur}) & POS, NER, NLI  \\
 Africaans  (\textsc{af}) & POS, NER      \\
 Estonian   (\textsc{et}) & POS, NER      \\
 Basque     (\textsc{eu}) & POS, NER      \\
 Farsi      (\textsc{fa}) & POS, NER      \\
 Hebrew     (\textsc{he}) & POS, NER      \\
 Hungarian  (\textsc{hu}) & POS, NER      \\
 Italian    (\textsc{it}) & POS, NER      \\
 Japanese   (\textsc{ja}) & POS, NER      \\
 Korean     (\textsc{ko}) & POS, NER      \\
 Marathi    (\textsc{mr}) & POS, NER      \\
 Portuguese (\textsc{pt}) & POS, NER      \\
 Vietnamese (\textsc{vi}) & POS, NER      \\
 Tamil      (\textsc{ta}) & POS, NER      \\
 Telugu     (\textsc{te}) & POS, NER      \\
 Swahili    (\textsc{sw}) & NLI      \\
 Thai       (\textsc{th}) & NLI      \\
\bottomrule
\end{tabular}
\caption{\label{tab:langs}
Languages used in our experiments.} 
\end{table}

\paragraph{Models and Hyperparameters}
Following \citet{conneau2020xlmr}'s \abr{xlm-r} Base model, we train transformers~\cite{vaswani2017transformer} with 12 layers, 768 units, 12 attention heads, and a maximum of 512 tokens per sequence.
To accommodate all languages and facilitate comparability between all pretraining setups, we use  \abr{xlm-r}'s vocabulary and the SentencePiece~\citep{kudo2018sentencepiece} tokenizer by~\citet{conneau2020xlmr}.


We use masked language modeling (\abr{mlm}) as our pretraining objective and, like \citet{devlin2019bert}, mask $15\%$ of the tokens.
We pretrain all models for 150K steps, using Adam W~\citep{adamw} with a learning rate of \num{1e-4} and a batch size of two on either NVIDIA RTX2080Ti or GTX1080Ti 12GB, on which it approximately took four days to train each model.
When pretraining, we preprocess sentences together to generate sequences of approximately 512 tokens.
For continued pretraining, we use a learning rate of \num{2e-5} and train for forty epochs, otherwise following the setup for pretraining.
For finetuning, we use a learning rate of \num{2e-5} and train for an additional ten epochs for both \abr{pos} tagging and \abr{ner}, and an additional five epochs for \abr{nli}, following \citet{hu20icml}.

\paragraph{Languages}
Table~\ref{tab:langs} shows the languages used in our experiments.
English is part of the pretraining data of all models. It is also the finetuning source language for all tasks, following \citet{hu20icml}.
We use two different sets of pretraining languages: ``Diverse (Div)'' and ``Related (Rel)'' (Table~\ref{tab:langs_pretrain}). 
We mainly focus on pretraining on up to five languages, except for \abr{pos} tagging where the trend is not clear and we further experiment on up to ten.


\begin{table}[t!]
\centering
\small
\setlength{\tabcolsep}{4pt}
\begin{tabular}{llllllllll}
\toprule
\bf Model & \bf Pretraining Languages\\
\midrule
Div-2 & \textsc{en}, \textsc{ru} \\
Div-3 & \textsc{en}, \textsc{ru}, \textsc{zh}\\
Div-4 & \textsc{en}, \textsc{ru}, \textsc{zh}, \textsc{ar}\\
Div-5 & \textsc{en}, \textsc{ru}, \textsc{zh}, \textsc{ar}, \textsc{hi}\\
Div-6 & \textsc{en}, \textsc{ru}, \textsc{zh}, \textsc{ar}, \textsc{hi}, \textsc{es} \\
Div-7 & \textsc{en}, \textsc{ru}, \textsc{zh}, \textsc{ar}, \textsc{hi}, \textsc{es}, \textsc{el} \\
Div-8 & \textsc{en}, \textsc{ru}, \textsc{zh}, \textsc{ar}, \textsc{hi}, \textsc{es}, \textsc{el}, \textsc{fi} \\
Div-9 & \textsc{en}, \textsc{ru}, \textsc{zh}, \textsc{ar}, \textsc{hi}, \textsc{es}, \textsc{el}, \textsc{fi}, \textsc{id}\\
Div-10 & \textsc{en}, \textsc{ru}, \textsc{zh}, \textsc{ar}, \textsc{hi}, \textsc{es}, \textsc{el}, \textsc{fi}, \textsc{id}, \textsc{tr}\\
\midrule
Rel-2 & \textsc{en}, \textsc{de}\\
Rel-3 & \textsc{en}, \textsc{de}, \textsc{sv} \\
Rel-4 &  \textsc{en}, \textsc{de}, \textsc{sv}, \textsc{nl}\\
Rel-5 &  \textsc{en}, \textsc{de}, \textsc{sv}, \textsc{nl}, \textsc{da}\\
\bottomrule
\end{tabular}
\caption{\label{tab:langs_pretrain}
Pretraining languages used for the models in our experiments: models are trained on  
a diverse set (Div-X) and related pretraining languages (Rel-X), with different numbers of pretraining languages.
}
\end{table}

For \abr{pos} tagging and \abr{ner}, we regard seventeen of the twenty-nine languages available in \abr{xtreme} as {\it unseen}, while the remaining twelve languages are pretraining languages for at least one model.
For \abr{nli}, six languages are {\it seen} and the rest are {\it unseen}. 
The order in which we add pretraining languages follows the size of their original \conll{} 2017 Wikipedia dumps, with larger sizes being added first.

\begin{figure}[t]
    \centering
    \includegraphics[width=0.45\textwidth]{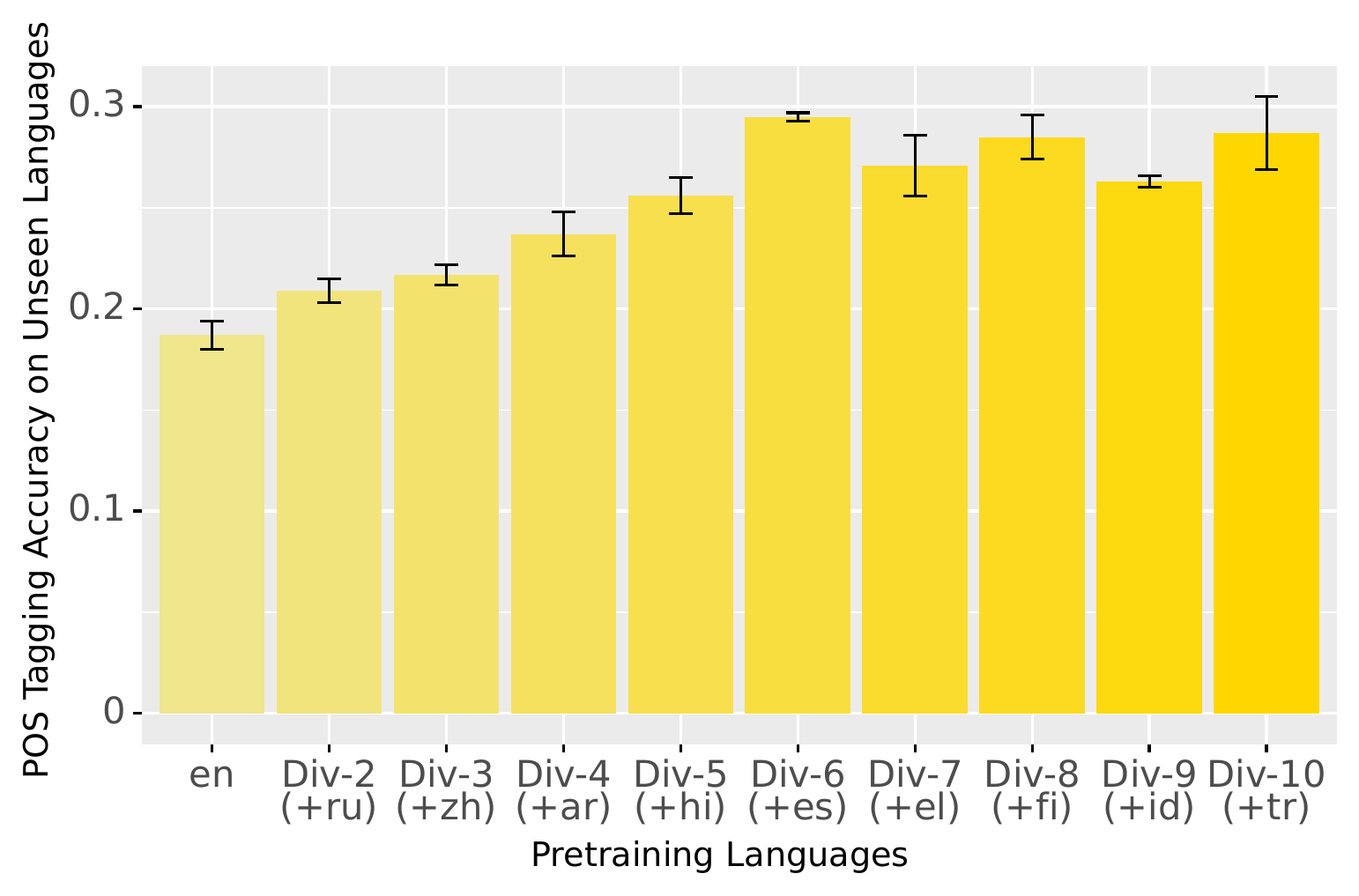}
    \caption{
    \abr{pos} tagging accuracy after pretraining on a diverse set of up to 10 languages and finetuning on English.
    The accuracy improves until six languages on the given target languages.
    }
    \label{fig:pos_unseen_avg}
\end{figure}

\begin{figure*}[th]
    \centering
    \includegraphics[width=0.99\textwidth]{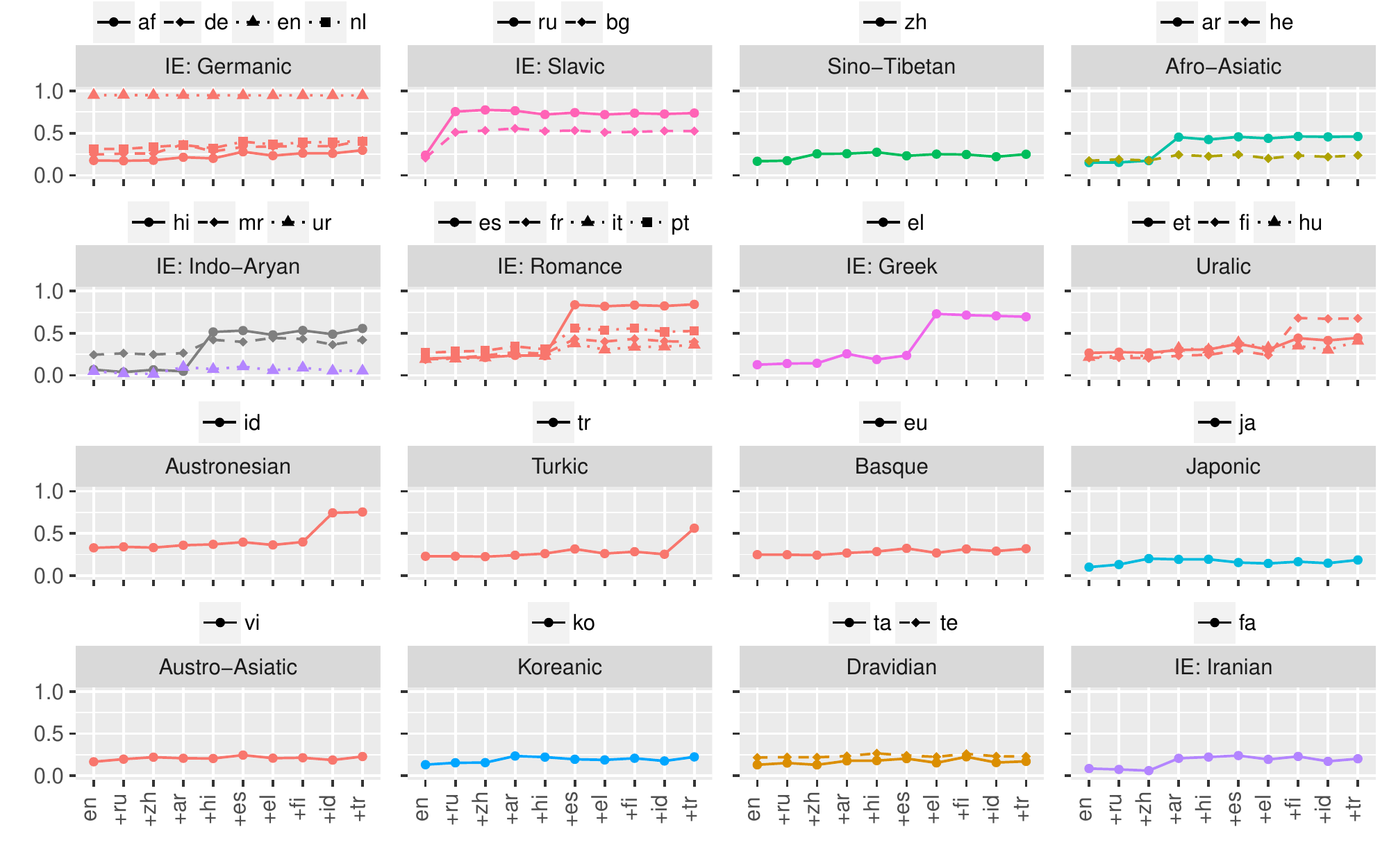}
    \caption{
    \abr{pos} tagging accuracy using models pretrained on a diverse set of languages (\textsc{en, ru, zh, ar, hi, es, el, fi, id, tr}) grouped by families of target languages, with Indo-European (\abr{ie}) languages further divided into subgroups following \abr{xtreme}.
    The colors represent the script type of the languages.
    The accuracy gain is larger when a pretraining language from the same family or using the same script is added.
    }
    \jbgcomment{Would be good if figure somehow showed that hte coloring was the same script.  Also, the shading seems to be filled in after the first shared script, but shouldn't it just be if the scripts match for that specific language?  Is there also a way to highlight when the languages actually match?  e.g., the rise for mulgarian on russian is far more interesting than the rise for russian on russian.}
    \yfcomment{The circles filled in after seeing the first shared script, that language is also used in the pretraining languages followed by. I believe it's better to keep it filled.}
    \label{fig:pos_unseen}
\end{figure*}

\section{Results}
\label{sec:results}

\jbgcomment{Don't like jumping directly from section into subsection}
We now present experimental results for each \abr{RQ}.

\subsection{Findings for RQ1}
\label{sec:more_lang_better}

\paragraph{POS Tagging}
Figure~\ref{fig:pos_unseen_avg} shows the \abr{pos} tagging accuracy averaged over the 17 languages unseen during pretraining.
On average, models pretrained on multiple languages have higher accuracy on unseen languages than the model pretrained exclusively on English, showing that the model benefits from a more diverse set of pretraining data. 
However, the average accuracy only increases up to six languages. This indicates that our initial hypothesis "the more languages the better" might not be true. 

Figure~\ref{fig:pos_unseen} provides a more detailed picture, showing the accuracy for different numbers of pretraining languages for all seen and unseen target languages.
As expected, accuracy jumps when a language itself is added as a pretraining language. 
Furthermore, accuracy rises if a 
pretraining language from the same language family as a target language is added:
for example, the accuracy of Marathi goes up by $9.3\%$ after adding Hindi during pretraining, and the accuracy of Bulgarian increases by $31.2\%$ after adding Russian.
This shows that related languages are indeed beneficial for transfer learning.
Also, (partially) sharing the same script with a pretraining language (e.g., \abr{es} and \abr{et}, \abr{ar} and \abr{fa}) helps with zero-shot cross-lingual transfer even for languages which are not from the same family.
These results are consistent with the outcome of~\citet{muller2020when} and partially support the hypothesis by~\citet{pires2019multilingual} that shared scripts are effective on unseen languages.

\jbgcomment{The linear regression could be better introduced; why are we doing it, what's the motivation, etc.}

But how important are the scripts compared to other features? 
To quantify the importance of it, we conduct a linear regression analysis on the \abr{pos} tagging result. 
Table~\ref{tab:pos_regression} shows the linear regression analysis results using typological features among target and pretraining languages.
For the script and family features, we follow \citet{xu2019crosslingual} and encoded them into binary values set to one if a language with the same script or from the same family is included as one of the pretraining languages. 
For syntax and phonology features, we derive those vectors from the URIEL database using lang2vec~\citep{littell2017uriel} following \citet{lauscher2020zero}.
We take the maximum cosine similarity between the target language and any of the pretraining languages.
Table~\ref{tab:pos_regression} further confirms that having a pretraining language which shares the same script contributes the most to positive cross-lingual transfer.

\begin{table}[t]
\centering
\small
\begin{tabular}{lrrc}
\toprule
\textbf{Features} & \textbf{Coef.} & \textbf{p-value} & \textbf{CI}\\
\midrule
Script            & \bf.061        & $<.001$          & [.050, .073]\\
Family            & .022           & $.004$           & [.007, .036]\\
Syntax            & .001           & $.905$           & [-.016, .018]\\
Phonology         & .021           & $<.001$          & [.009, .033]\\
\# pretrain langs & .011           & $.044$           & [.000, .022]\\
\bottomrule
\end{tabular}
\caption{\label{tab:pos_regression}
Regression analysis on the \abr{pos} tagging accuracy with coefficients (Coef.), p-value, and $95\%$ confidence interval (CI). 
A large coefficient with a low p-value indicates that the feature significantly contributes to better cross-lingual transfer, which shows that the same script is the most important feature.
}
\end{table}

We sadly cannot give a definitive optimal number of pretraining languages. 
One consistent finding is that, for the large majority of languages, using only English yields the worst results for unseen languages. However, adding pretraining languages does not necessarily improve accuracy (Figure \ref{fig:pos_unseen_avg}). This indicates that, while we want more than one pretraining language, using a smaller number than the 100 commonly used pretraining languages is likely sufficient 
unless we expect them to be closely related to one of the potential target languages.

\paragraph{NER}
Our \abr{ner} results show a similar trend. Therefore, we only report the average performance in the main part of this paper (Figure~\ref{fig:ner_unseen_avg}), and full details are available in Appendix~\ref{sec:appendix_ner}.
For \abr{ner}, transfer to unseen languages is more limited, likely due to the small subset of tokens which are labeled as entities when compared to \abr{pos} tags.

\begin{figure}[t]
    \centering
    \includegraphics[width=0.45\textwidth]{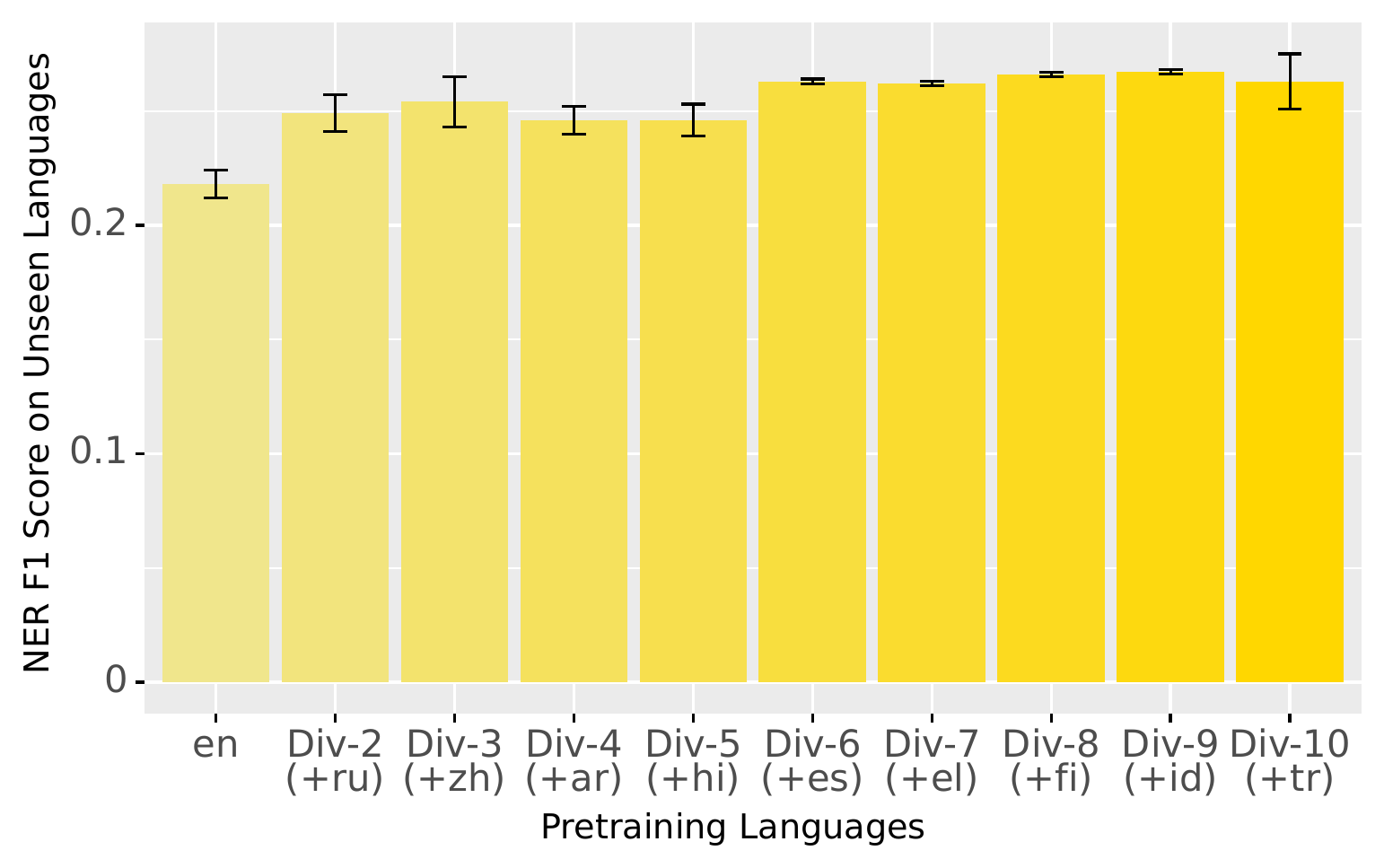}
    \caption{
    \abr{ner} F1 score after pretraining on a diverse set of up to 10 languages and finetuning on English.
    }
    \label{fig:ner_unseen_avg}
\end{figure}

\begin{figure}[t]
    \centering
    \includegraphics[width=0.45\textwidth]{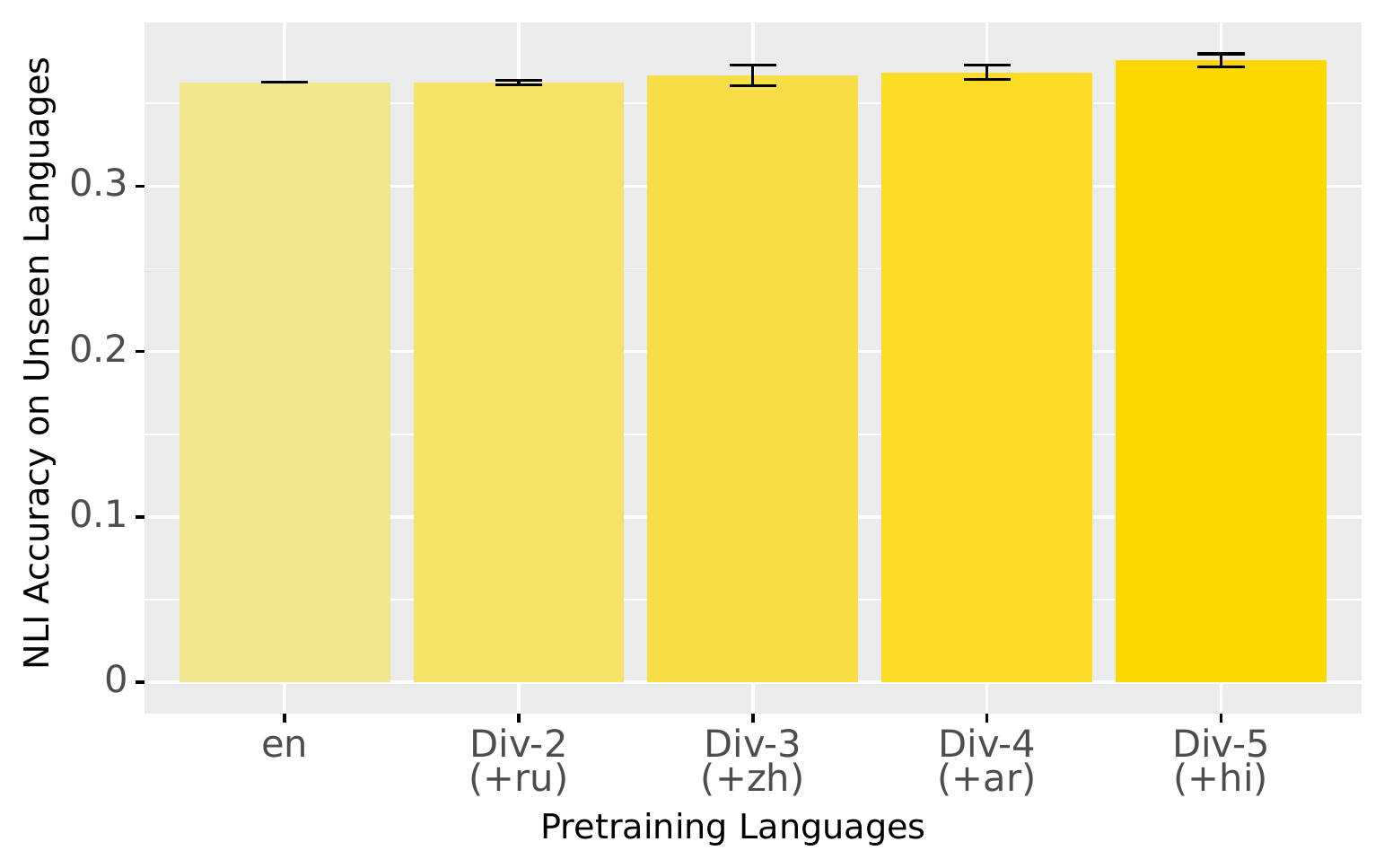}
    \caption{
    \abr{xnli} accuracy after pretraining on a diverse set and finetuning on English.
    }
    \label{fig:xnli_unseen_diverse}
\end{figure}

\paragraph{NLI}
Our \abr{nli} results in Figure~\ref{fig:xnli_unseen_diverse} show a similar trend: 
accuracy on unseen languages plateaus at a relatively small number of pretraining languages. Specifically, Div-4 has the highest accuracy for 8 target languages, while Div-5 is best only for two target languages.
Accuracy again increases with related languages, such as an improvement of \mbox{$3.7\%$} accuracy for Bulgarian after adding Russian as a pretraining language. 
Full results are available in Appendix~\ref{sec:appendix_nli}.

\begin{figure}[t]
    \centering
    \begin{subfigure}[b]{0.495\textwidth}
      \centering
      \includegraphics[width=0.95\textwidth]{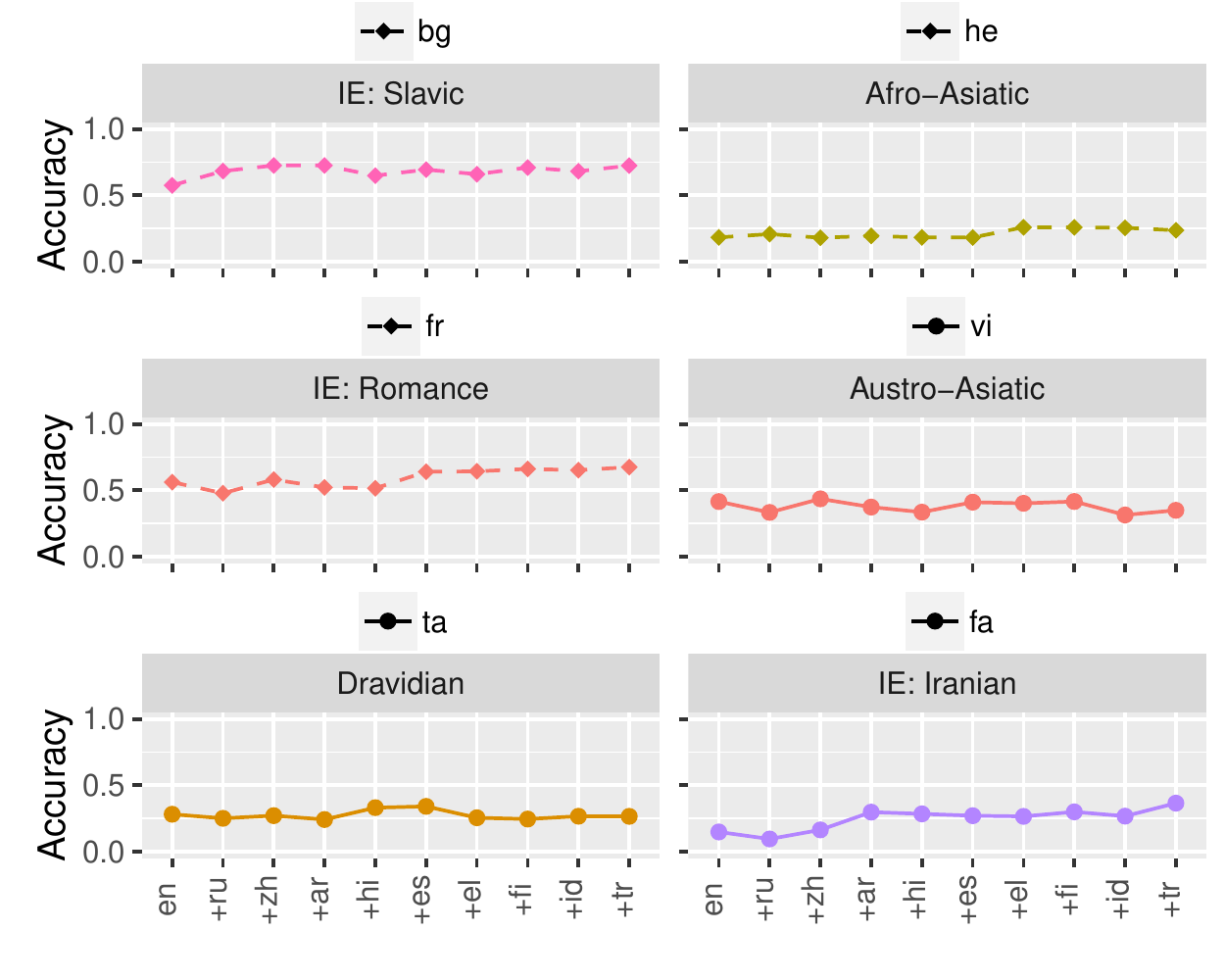}
      \caption{
      \abr{pos} tagging accuracy.
      }
      \label{fig:pos_unseen_continued_el}
    \end{subfigure}
    \begin{subfigure}[b]{0.495\textwidth}
      \centering
      \includegraphics[width=0.95\textwidth]{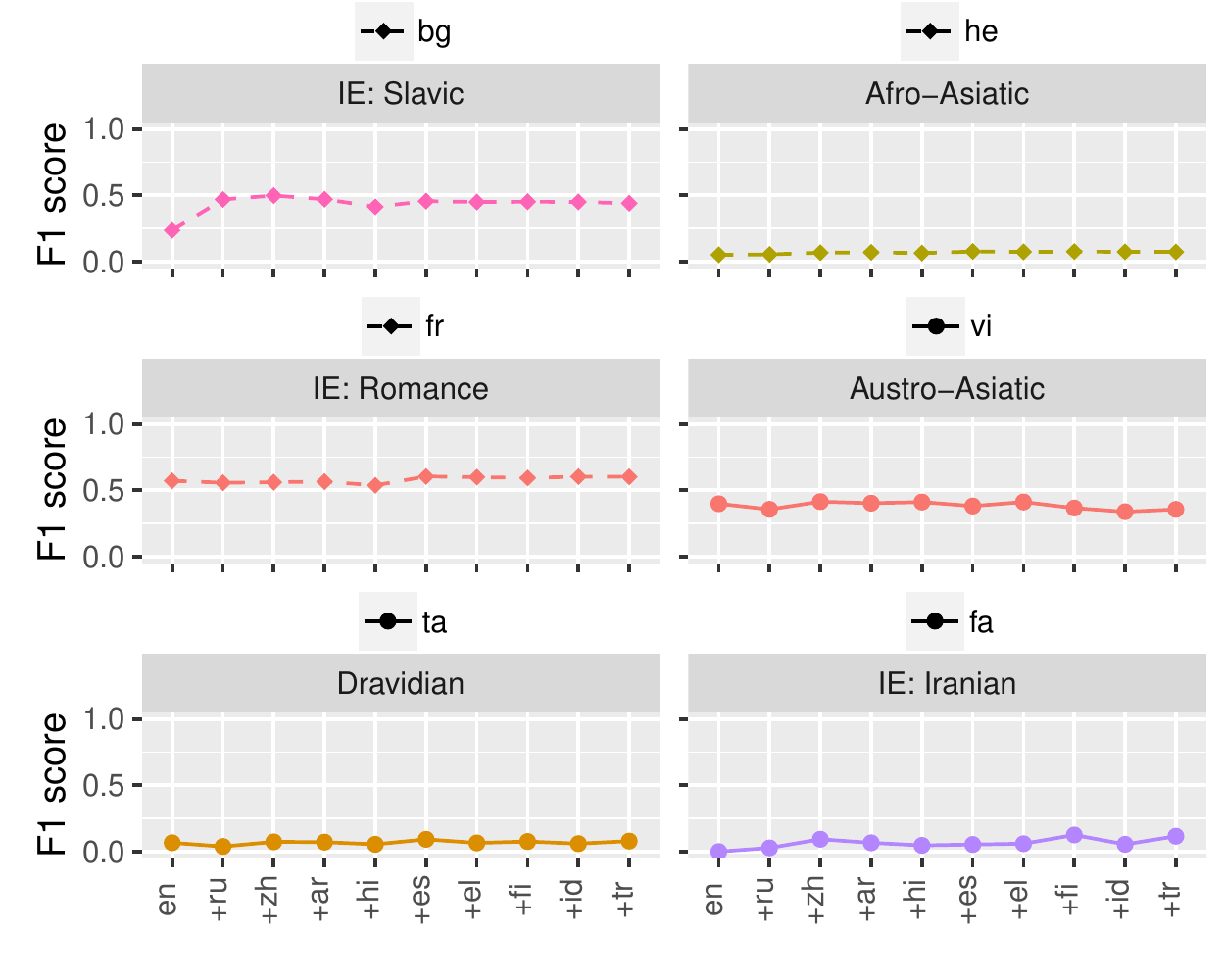}
      \caption{
      \abr{ner} F1 scores.
      }
      \label{fig:ner_unseen_continued}
    \end{subfigure}
    \caption{
    Results after continued training on the Bible of each target language.
    The continued training gives limited improvement on \abr{ner} for most languages when compared to \abr{pos} tagging.}
\end{figure}

\subsection{Findings for RQ2}
\paragraph{POS Tagging} Figure~\ref{fig:pos_unseen_continued_el} shows the \abr{pos} tagging results for six languages after adaptation of the pretrained models via continued pretraining.
As expected, accuracy is overall higher than in Figure~\ref{fig:pos_unseen}.
Importantly, 
there are accuracy gains in Farsi when adding Turkish ($+9.8\%$) and in Hebrew when adding Greek ($+7.7\%$), which are not observed before adapting models.
We further investigate it in Section~\ref{sec:xlmr}.


\paragraph{NER} 
\abr{ner} results in Figure~\ref{fig:ner_unseen_continued} show similarities between \abr{pos} tagging (e.g., improvement on Bulgarian after adding Russian).
However, there is limited improvement on Farsi after adding Arabic despite partially shared scripts between the two languages. This indicates that the effect of adding related pretraining languages is partially task-dependent. 

\begin{figure*}[t]
    \centering
    \includegraphics[width=0.95\textwidth]{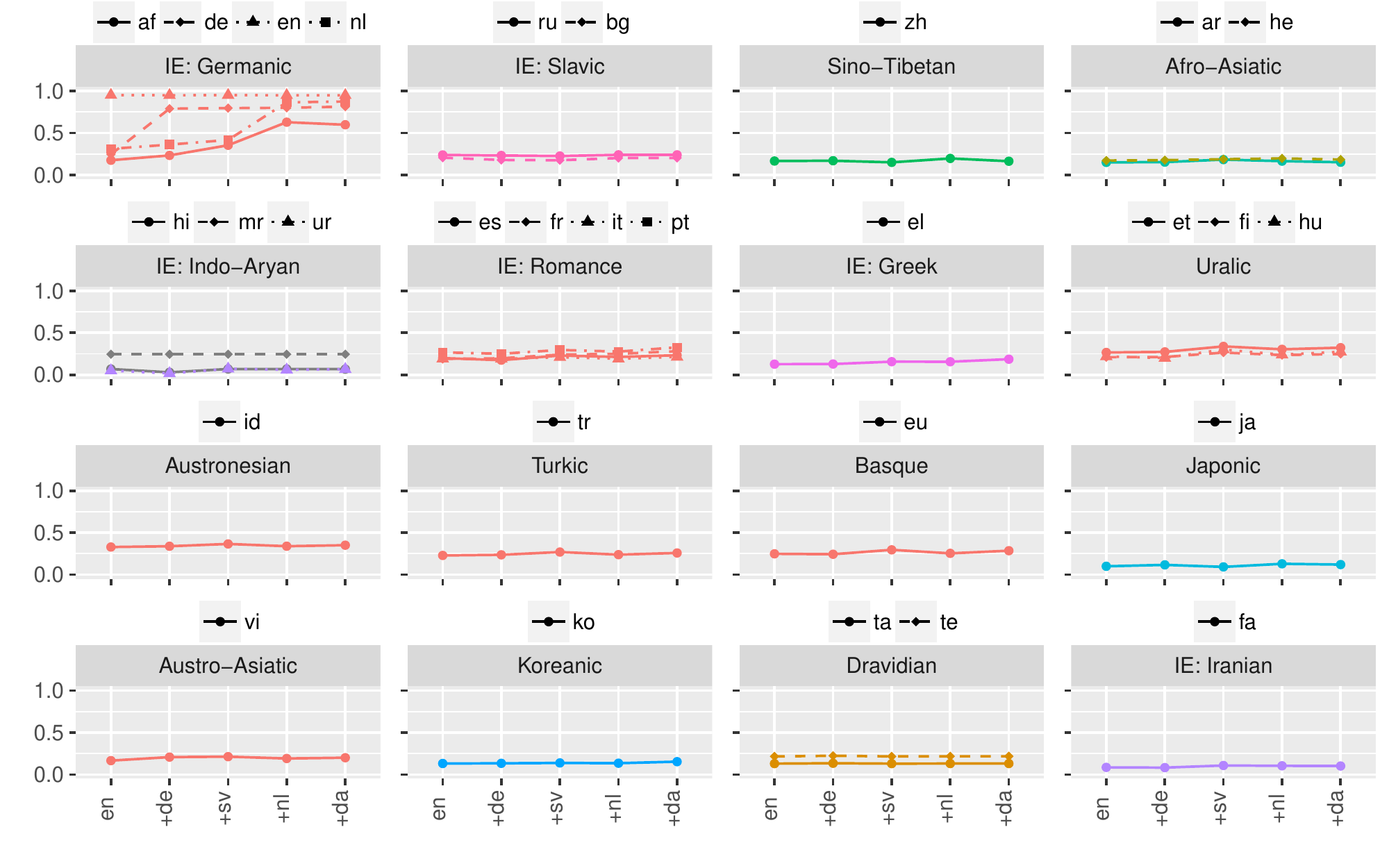}
    \caption{\abr{pos} tagging accuracy using related pretraining languages (\textsc{en, de, sv, nl, da}) grouped by families of target languages, with Indo-European (\abr{ie}) languages further divided into subgroups following the \abr{xtreme} dataset.
    A change in accuracy can mainly be observed for Germanic, Romance, and Uralic languages due to only using pretraining languages from the Germanic language family. 
    }
    \label{fig:pos_unseen_related}
\end{figure*}
\begin{figure}[t]
    \centering
    \includegraphics[width=0.45\textwidth]{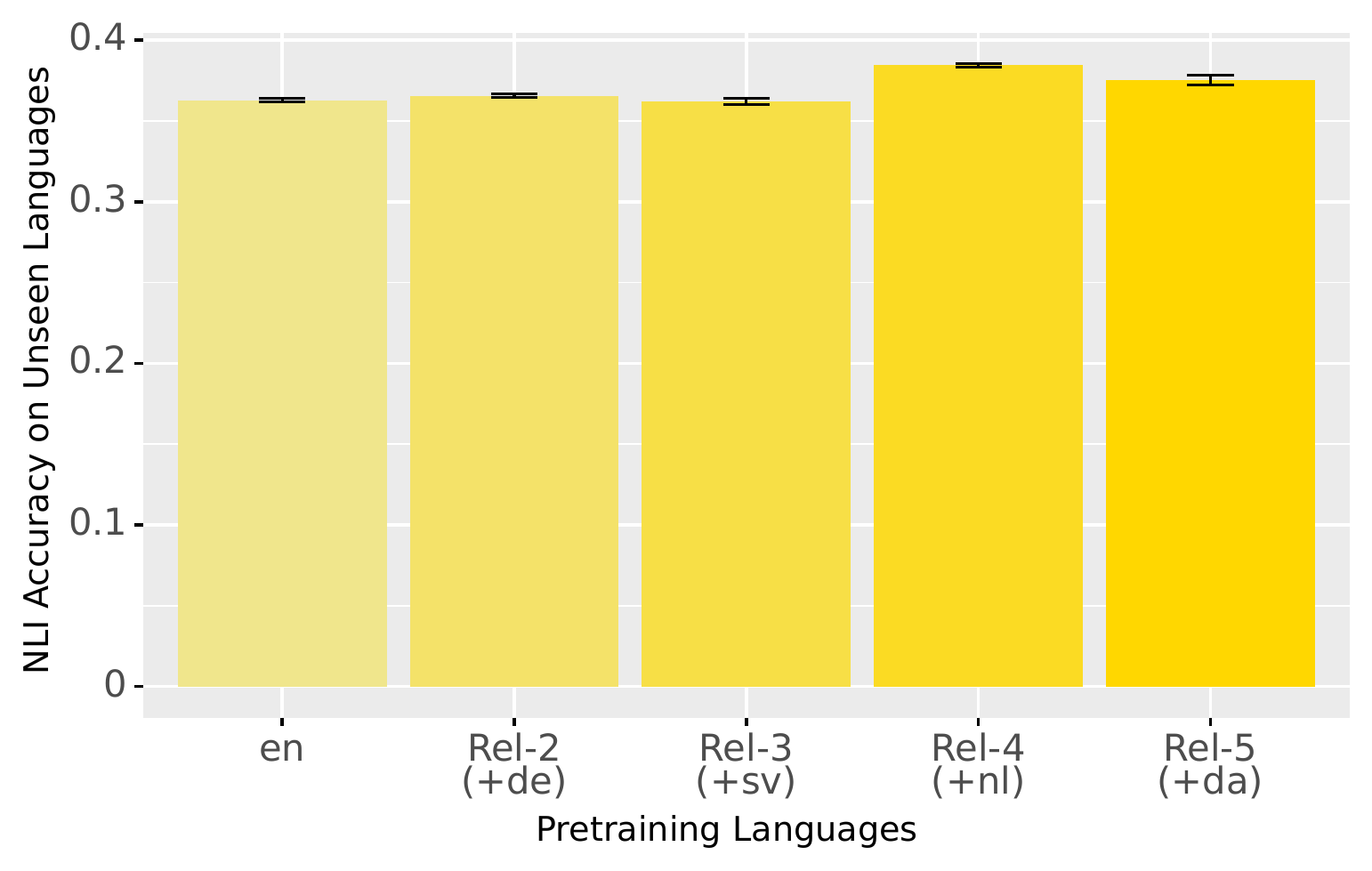}
    \caption{
    \abr{xnli} accuracy on 10 unseen languages after pretraining on a set of related languages and finetuning on English.
    }
    \label{fig:xnli_unseen_related}
\end{figure}

\paragraph{NLI} 
For \abr{nli}, accuracy increases slightly after adding a second pretraining language. 
Results for two to five pretraining languages are similar for all target languages and, for Greek and Turkish, still similar to the English-only model. This indicates that, similar to our findings for \abr{pos} tagging, a few pretraining languages could be sufficient for model adaptation. 
Full results are available in Appendix~\ref{sec:appendix_nli}.
Finally, our \abr{nli} results are low overall.
This is likely due to the size of the pretraining corpus being one of the top correlated features for \abr{nli}~\citep{lauscher2020zero}, unlike for \abr{pos} tagging~\citep{hu20icml}.

\subsection{Findings for RQ3}


\paragraph{POS Tagging}
In contrast to \abr{rq}1, \abr{pos} tagging accuracy changes for most languages are limited when increasing the number of pretraining languages (Figure~\ref{fig:pos_unseen_related}). 
The unseen languages on which we observe gains 
belong to the Germanic, Romance, and Uralic language families, which are relatively (as compared to the other language families) close to English.
The accuracy on languages from other language families changes by $<10\%$, which is smaller than the change for a diverse set of pretraining languages.
This indicates that the models pretrained on similar languages struggle to transfer to unrelated languages.

\paragraph{NER}
F1 scores of \abr{en}, Rel-2, Rel-3, Rel-4, and Rel-5 are .218, .219, .227, .236, and .237 respectively. 
Compared to Div-X, pretraining on related languages also improves up to adding five languages. However, these models bring a smaller improvement, similar to \abr{pos} tagging.

\paragraph{NLI} 
Figure~\ref{fig:xnli_unseen_related} shows a similar trend for \abr{nli}: when adding related pretraining languages, accuracy on languages far from English either does not change much or decreases. In fact, for nine out of thirteen unseen target languages, Rel-5 is the worst.



\begin{figure*}[th]
    \centering
    \begin{subfigure}[b]{0.495\textwidth}
         \centering
        \includegraphics[width=0.99\textwidth]{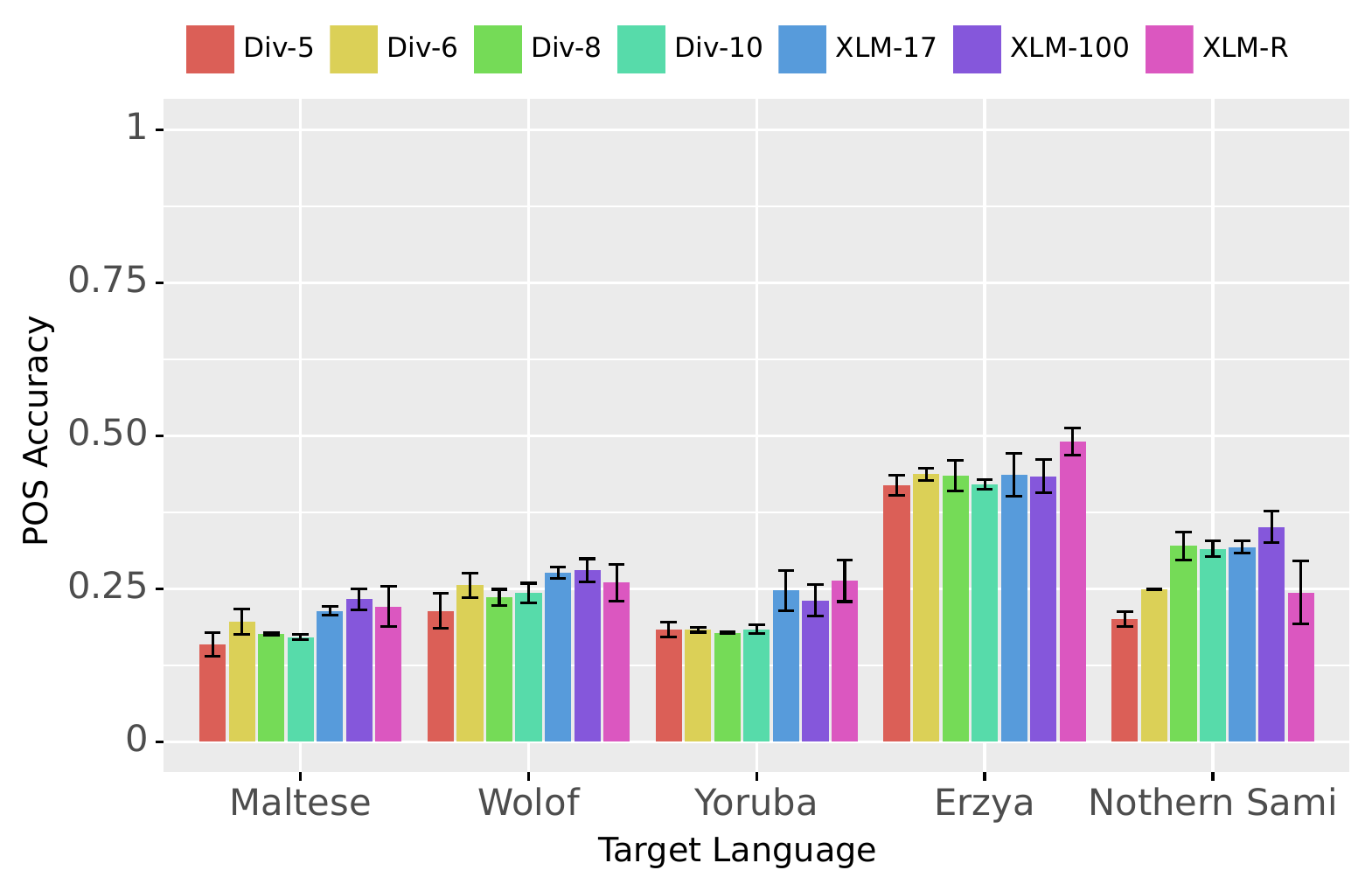}
            \caption{
            Before model adaptation. 
            }
            \label{fig:pos_lowresource_before_adapt}
    \end{subfigure}
    \hfill
    \begin{subfigure}[b]{0.495\textwidth}
        \centering
        \includegraphics[width=0.99\textwidth]{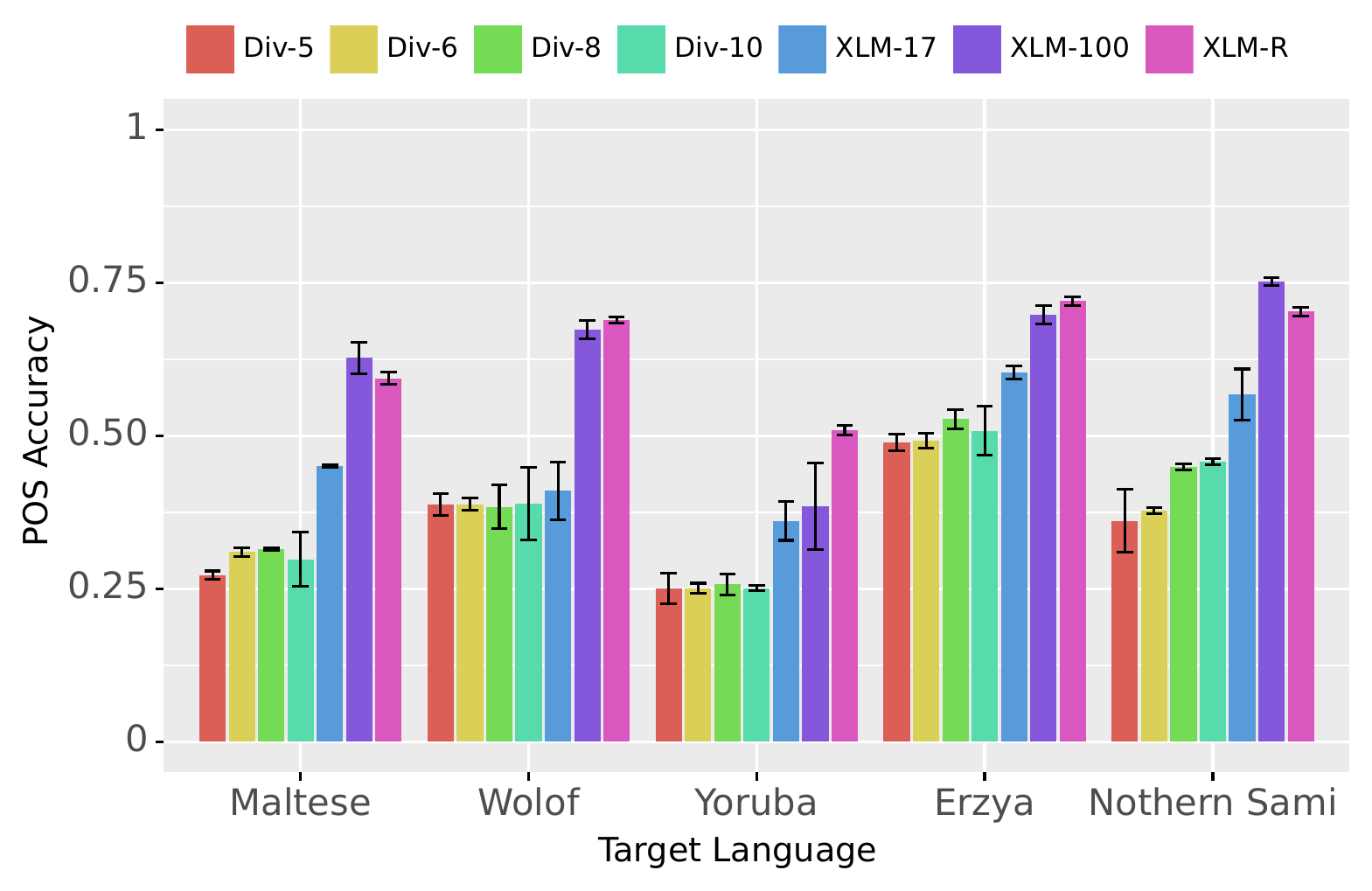}
            \caption{
            After model adaptation. 
            }
            \label{fig:pos_lowresource_after_adapt}
    \end{subfigure}
    \caption{
            \abr{pos} tagging accuracy of our models pretrained on a diverse set of languages, \abr{xlm}-17, \abr{xlm}-100, and \abr{xlm-r} after finetuning on English.
            The models before adaptation are roughly on par regardless of the number of pretraining languages, and the models after adaptation are more affected by related pretraining languages. 
            }
\jbgcomment{This figure looks a little different from the others: there's no spacing between bars.  Would be good to make this consistent.} 
\end{figure*}

\section{More Pretraining Languages}
\label{sec:xlmr}

Our main takeaways from the last section are: 
(\abr{rq}1) without model adaptation, increasing the number of pretraining languages does not improve accuracy on unrelated unseen target languages;
(\abr{rq}2) model adaptation largely helps exploiting models pretrained on more languages; and 
(\abr{rq}3) when using more than one pretraining language, diversity is important.

However, there are limitations in the experimental settings in Section~\ref{sec:results}.
We assume the following:
(1) relatively small pretraining corpora; 
(2) the target languages are included when building the model's vocabulary;
(3) fixed computational resources; and 
(4) only up to ten pretraining languages. 
We now explore if our findings for \abr{rq}1 and \abr{rq}2 hold without such limitations.
For this, we use two publicly available pretrained \abr{\abr{xlm}} models \citep{conneau2019xlm}, which have been pretrained on full size Wikipedia in 17 (\abr{xlm}-17) and 100 (\abr{xlm}-100) languages, and \abr{xlm-r} base model trained on a larger Common Crawl corpus~\citep{conneau2020xlmr} in 100 languages.
We conduct a case study on low-resource languages unseen for all  models, including unseen vocabularies: Maltese (\textsc{mt}), Wolof (\textsc{wo}), Yoruba (\textsc{yo}), Erzya (\textsc{myv}), and Northern Sami (\textsc{sme}).
All pretraining languages used in Div-X are included in \abr{xlm}-17 except for Finnish, and all 17 pretraining languages for \abr{xlm}-17 are a subset of the pretraining languages for \abr{xlm}-100.
We report the averages with standard deviations from three random seeds.


\subsection{Results} 

\paragraph{RQ1} 
For models without adaptation,
accuracy does not improve for increasing numbers of source languages (Figure~\ref{fig:pos_lowresource_before_adapt}). 
Indeed, the accuracy on both \abr{xlm}-17 and \abr{xlm}-100 are on par even though the former uses 17 pretraining languages and the latter uses 100. 
One exception is Northern Sami (Uralic language with Latin script)
due to \abr{xlm}-17 not seeing any Uralic languages, but \abr{xlm}-100 does during pretraining.
When further comparing Div-10 and \abr{xlm}-17, increase in accuracy by  additional pretraining languages is limited.
Erzya remains constant from five to 100 languages (except for \abr{xlm-r}), even when increasing the pretraining corpus size from downsampled (Div-X) to full Wikipedia (\abr{xlm}-17 and \abr{xlm}-100).



\paragraph{RQ2} 
For the models with adaptation (Figure~\ref{fig:pos_lowresource_after_adapt}),
there is a significant gap between \abr{xlm}-17 and \abr{xlm}-100.
This confirms our findings in the last section: more pretraining languages is beneficial if the pretrained models are adapted to the target languages. 
Thus, a possible explanation is that one or more of \abr{xlm}-100's pretraining languages is similar to our target languages and such languages can only be exploited through continued pretraining (e.g., Ukrainian included in \abr{xlm}-100 but not in Div-X).
Therefore, having the model see more languages during pretraining is better when the models can be adapted to each target language.

\section{Related Work}

\paragraph{Static Cross-lingual Word Embeddings}
Static cross-lingual word embeddings \citep{mikolov2013exploit,lample2018word} embed and align words from multiple languages for downstream NLP tasks~\citep{lample2018unsupervised,D18-1398}, including a massive one trained on 50+ languages~\cite{ammar2016massively}.
Static cross-lingual embedding methods can be classified into two groups: supervised and unsupervised. 
Supervised methods use bilingual lexica as the cross-lingual supervision signal. On the other hand,  pretrained multilingual language models and  unsupervised cross-lingual embeddings are similar because they do not use a bilingual lexicon.
\citet{lin2019choosing} explore the selection of transfer language using both data-independent (e.g., typological) features, and data-dependent features (e.g., lexical overlap). 
Their work is on static supervised cross-lingual word embeddings, whereas this paper explores pretrained language models.

\jbgcomment{I think you can talk more in depth about word embeddings: contrast the supervision of the approaches and compare the massively multilingual word embeddings}

\yfcomment{TODO: Mention that Bilingual lexicons are needed for supervised cross-lingual word embeddings, pretrained models are closer to unsupervised cross-lingual word embeddings}

\paragraph{Analysis of Pretrained Multilingual Models on Seen Languages}
Starting from  \citet{pires2019multilingual}, analysis of the cross-lingual transferability of pretrained multilingual language models has been a topic of interest. 
\citet{pires2019multilingual} hypothesize that cross-lingual transfer occurs due to shared tokens across languages, but  \citet{artetxe2020crosslingual} show that cross-lingual transfer can be successful even among languages without shared scripts.
Other work 
investigates the relationship between zero-shot cross-lingual learning and typological features \citep{lauscher2020zero}, encoding language-specific features~\citep{libovicky2020language}, and 
\mbert{}'s multilinguality~\citep{dufter2020identifying}.
However, the majority of analyses have either been limited to large public models (e.g., \mbert{}, \abr{xlm-r}), to up to two pretraining  languages \citep{k2020crosslingual,wu2020languages}, or to target languages seen during pretraining. 
One exception is the concurrent work by~\citet{devries2022when} on analyzing the choice of language for the task-specific training data on unseen languages.
Here, we analyze the ability of models to benefit from an increasing number of pretraining languages.

\section{Conclusion}
This paper explores the effect which pretraining on different numbers of languages has on unseen target languages after finetuning on English. We find: 
(1) if not adapting the pretrained multilingual language models to target languages, a set of diverse pretraining languages which covers the script and family of unseen target languages (e.g., 17 languages used for \abr{xlm}-17) is likely sufficient; and
(2) if adapting the pretrained multilingual language model to target languages, then one should pretrain on as many languages as possible up to at least 100.

Future directions include analyzing the effect of multilingual pretraining from different perspectives such as different pretraining tasks and architectures, e.g., mT5~\citep{xue2021mt5}, and more complex tasks beyond classification or sequence tagging.

\jbgcomment{Is it possible to make these suggestions firmer?  E.g., tell people how many languages you need, what constitutes diversity, etc.?  If we want this paper to get cited, we need a rulebook that people should follow for multilingual LMs.}

\yfcomment{Should be based on the regression analysis.}




\section*{Acknowledgements}
We sincerely thank the reviewers for their constructive and detailed feedback. 
We also thank the members of University of Colorado Boulder's NALA group, especially
Abteen Ebrahimi for providing the code and St{\'e}phane Aroca-Ouellette for giving feedback on an early draft.  Boyd-Graber is supported by ODNI, IARPA,
via the BETTER Program contract \#2019-19051600005. The views and conclusions contained herein
are those of the authors and should not be interpreted as necessarily representing the official policies,
either expressed or implied, of ODNI, IARPA, or the U.S. Government. The U.S. Government
is authorized to reproduce and distribute reprints for governmental purposes notwithstanding any
copyright annotation therein.
\bibliography{bib/journal-full,bib/yoshinari,bib/custom}
\bibliographystyle{acl_natbib}
\vspace*{\fill} 
\pagebreak
\appendix
\clearpage 

\section{\abr{ner} Results}
\label{sec:appendix_ner}

\begin{figure}[t]
    \centering
    \includegraphics[width=0.49\textwidth]{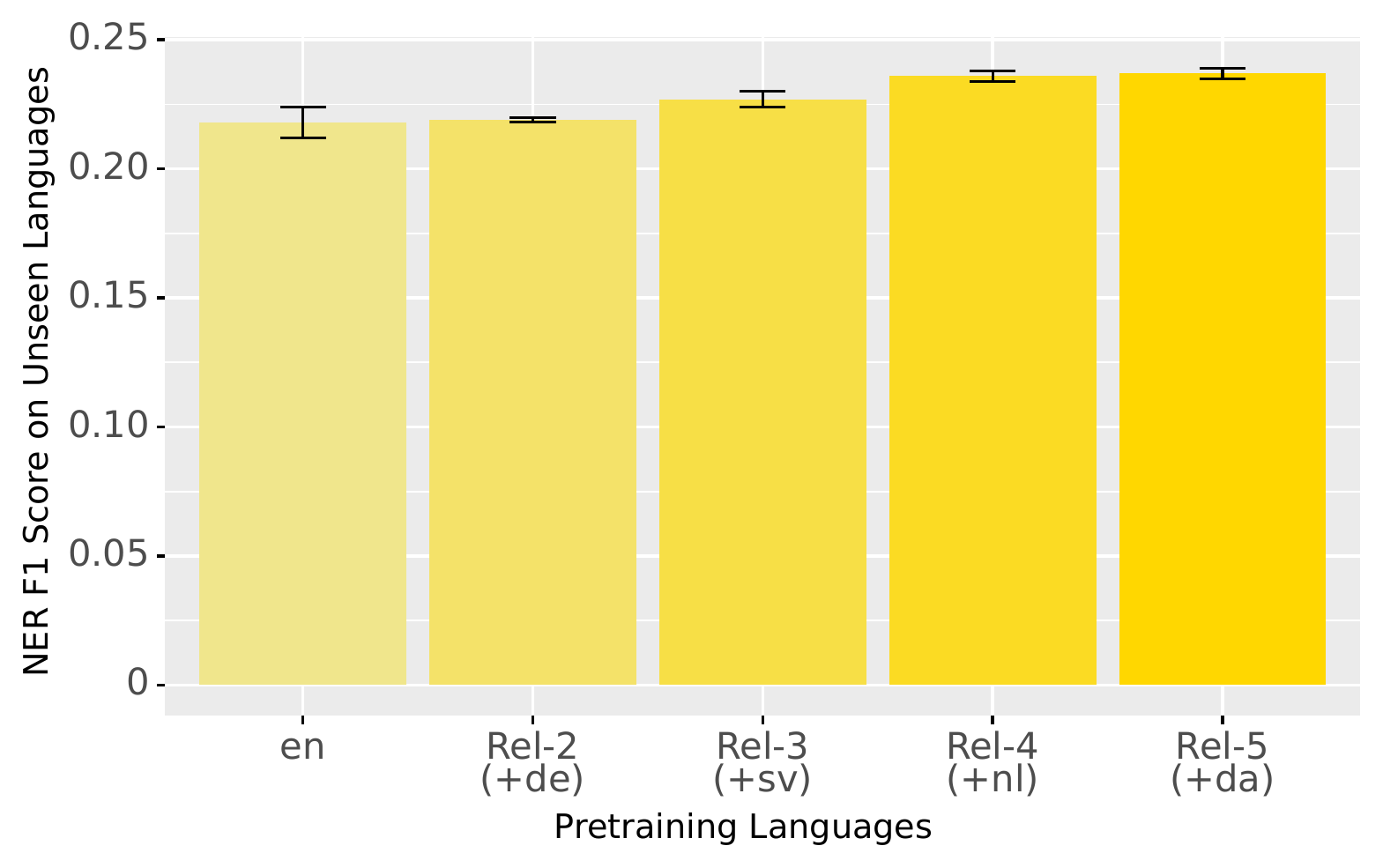}
    \caption{\abr{ner} F1 score using related pretraining languages (\textsc{en, de, sv, nl, da})
    }
    \label{fig:ner_unseen_related_avg}
\end{figure}

\begin{figure*}[t]
    \centering
    \includegraphics[width=0.99\textwidth]{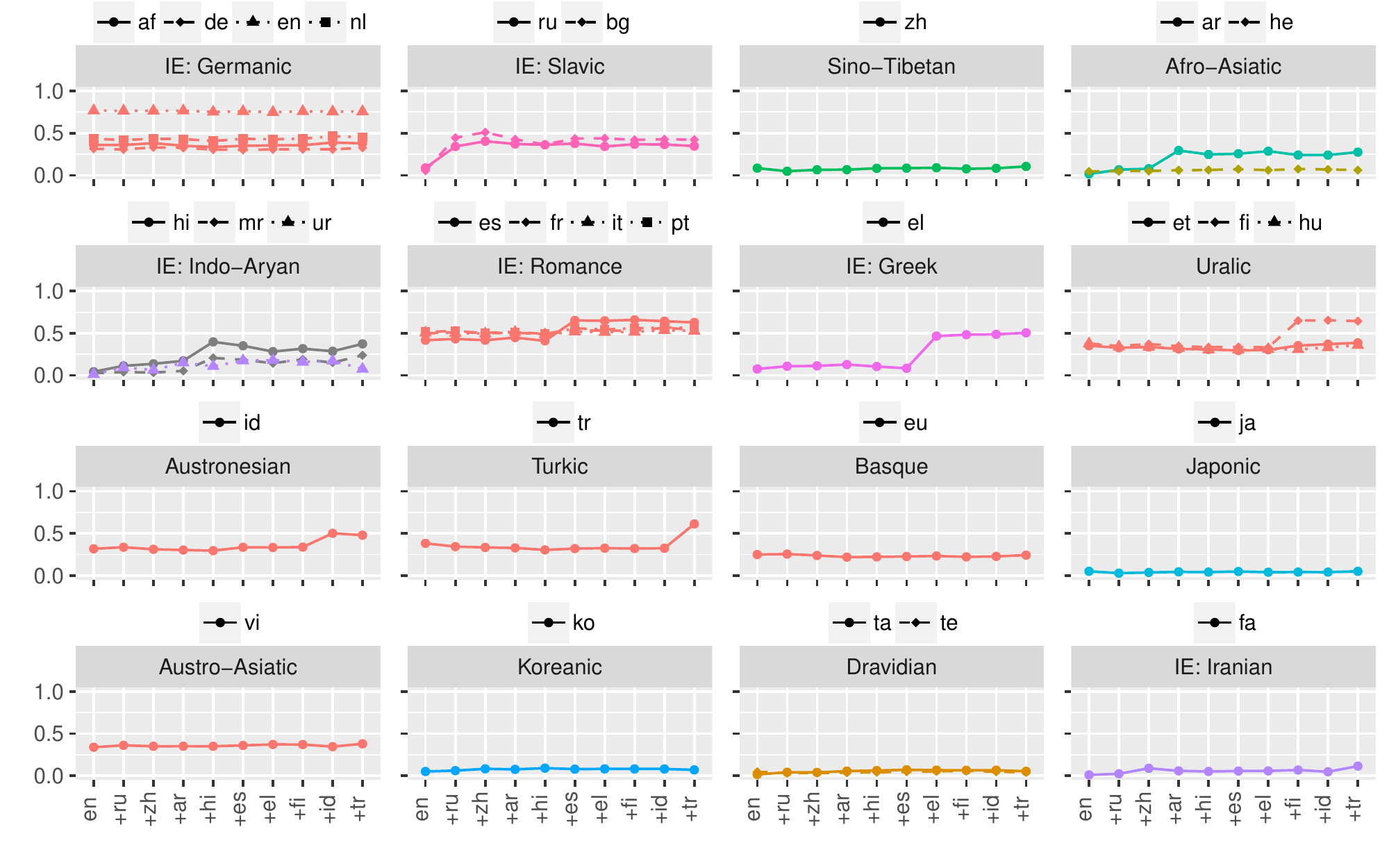}
    \caption{
        \abr{ner} F1 score on diverse pretraining languages (\textsc{en, ru, zh, ar, hi, es, el, fi, id, tr}) grouped by families of target languages, with Indo-European (\abr{ie}) languages further divided into subgroups following \abr{xtreme}.
        The accuracy gain is significant for seen pretraining languages, and also the languages from the same family of the pretraining languages when added. 
    }
    \label{fig:ner_unseen}
\end{figure*}

We show additional experimental results on \abr{ner} in Figures~\ref{fig:ner_unseen_related_avg} and \ref{fig:ner_unseen}.

\begin{table}[t]
\centering
\small
\setlength{\tabcolsep}{8pt}
\begin{tabular}{lrrrr}
\toprule
\textbf{Pretrain} & \textsc{el} & \textsc{vi} &  \textsc{tr} & \textsc{fr} \\
\midrule
\textsc{en}             & .351 & .367 & .365 & .395 \\
Div-2 (+ru)               & .360 & .411 & .372 & .436 \\
Div-3 (+zh)               & .353 & .386 & .368 & .403 \\
Div-4 (+ar)               & .362 & .395 & .374 & .438 \\
Div-5 (+hi)               & .358 & .389 & .376 & .418 \\
\bottomrule
\end{tabular}
\caption{\label{tab:xnli_continued}
\abr{nli} accuracy after pretraining on a diverse set of up to 5 languages, continued pretraining on the target-language Bible, and finetuning on English.
}
\end{table}

\section{\abr{nli} Results}
\label{sec:appendix_nli}
Tables~\ref{tab:xnli_unseen} and \ref{tab:xnli_related} shows the results without model adaptation, and Table~\ref{tab:xnli_continued} shows the full results with model adaptation.

\begin{table*}[t]
\centering
\small
\setlength{\tabcolsep}{5pt}
\begin{tabular}{l|rrrrr|rrrrrrrrrr}
\toprule
\textbf{Pretrain} & en & ru & zh & ar & hi & bg & de & el & es & fr & sw & th & tr & ur & vi \\
\midrule
\textsc{en}           & .731 & .343 & .340 & .339 & .345 & .347 & .375 & .346 & .404 & .381 & .366 & .350 & .358 & .347 & .354 \\
Div-2        & .725 & .457 & .336 & .341 & .342 & .384 & .373 & .346 & .421 & .382 & .364 & .342 & .354 & .338 & .352 \\
Div-3     & .738 & .500 & .485 & .336 & .338 & .389 & .374 & .341 & .412 & .382 & .354 & .340 & .345 & .339 & .345 \\
Div-4    & .718 & .452 & .467 & .460 & .350 & .418 & .398 & .352 & .439 & .417 & .379 & .351 & .369 & .361 & .361 \\
Div-5    & .717 & .466 & .484 & .460 & .462 & .426 & .382 & .346 & .443 & .386 & .370 & .348 & .356 & .349 & .349\\
\bottomrule
\end{tabular}
\caption{\label{tab:xnli_unseen}
\abr{nli} accuracy on diverse pretraining languages over five seen (\textsc{en,ru,zh,ar,hi}) and 10 unseen languages.
}
\end{table*}

\begin{table*}[t]
\centering
\small
\setlength{\tabcolsep}{5pt}
\begin{tabular}{l|rr|rrrrrrrrrrrrr}
\toprule
\textbf{Pretrain} & en & de   & ru  & zh    & ar   & hi   & bg  & el   & es   & fr   & sw   & th   & tr   & ur   & vi   \\
\midrule                                                           
\textsc{en}   & .731        & .375 & .343 & .340 & .339 & .345 & .347 & .346 & .404 & .381 & .366 & .350 & .358 & .347 & .354 \\
Rel-2         & .733 & .536 & .363 & .350 & .357 & .361 & .359 & .367 & .422 & .384 & .374 & .360 & .381 & .363 & .369 \\
Rel-3         & .721        & .535 & .351 & .349 & .350 & .355 & .350 & .352 & .434 & .420 & .383 & .357 & .382 & .348 & .370 \\
Rel-4         & .710        & .493 & .350 & .336 & .348 & .355 & .354 & .349 & .433 & .409 & .368 & .360 & .373 & .347 & .363 \\
Rel-5         & .726        & .527 & .339 & .335 & .335 & .342 & .343 & .342 & .430 & .415 & .376 & .339 & .372 & .335 & .347 \\
\bottomrule
\end{tabular}
\caption{\label{tab:xnli_related}
\abr{nli} accuracy on the 13 unseen languages using the models pretrained on related languages (\textsc{en, de, sv, nl, da}), incrementally added one language at a time up to five languages.
}
\end{table*}

\section{Notes on the Experimental Setup for Model Adaptation}
Following are the additional notes on the setup of the model adaptation:
\begin{itemize}
 \item No vocabulary augmentation is conducted unlike~\citet{wang2020extending}. We use \abr{xlm-r}'s vocabulary throughout all experiments in this paper.
 \item The Bible is used instead of Wikipedia for the continued pretraining or model adaptation to minimize the corpus size and contents inconsistency across languages.
\end{itemize}

\end{document}